\documentclass[runningheads]{llncs}
\newif\ifanon \anonfalse
\usepackage[T1]{fontenc}
\usepackage{graphicx}
\usepackage{amsmath,amssymb}
\usepackage{booktabs}
\usepackage{siunitx}
\usepackage{subcaption}
\usepackage{url}
\usepackage[hidelinks]{hyperref}

\ifanon
  \newcommand{\modelname}{CDN}
  \newcommand{\repourl}{\url{https://anonymous.4open.science/r/cdn-dhcd}}
\else
  \newcommand{\modelname}{Barnamala}
  \newcommand{\repourl}{\url{https://github.com/Ampixa/barnamala}}
\fi

\begin{document}
\title{\modelname: Parameter-Efficient Handwritten Devanagari Recognition at Benchmark Saturation}
\titlerunning{\modelname: Parameter-Efficient Devanagari Recognition}
\ifanon
  \author{Anonymous}
  \authorrunning{Anonymous}
  \institute{Anonymous Institution}
\else
  \author{Ashish Thapa\inst{1} \and Samrat Karki\inst{2}}
  \authorrunning{A.~Thapa and S.~Karki}
  \institute{Ampixa Labs \\ \email{ashish@ampixa.com}
  \and
  Pulchowk Campus \\ \email{082bel077.samrat@pcampus.edu.np}}
\fi
\maketitle

\begin{abstract}
We built a compact convolutional network (\textbf{1.11\,M} parameters) for 46-class
DHCD Devanagari recognition and reached \textbf{99.73\,\%}~\cite{acharya2015dhcd},
the highest reported at \textbf{15.6\(\times\)} smaller than prior state-of-the-art.
We have effectively reached the saturation point: every model tested, large teacher
ensembles included, hits the same \textbf{11-error intrinsic floor}.
No configuration achieves a statistically clear win under exact McNemar tests
with Wilson confidence intervals. Even without knowledge distillation, our student
matches the nearest large-model baseline (17.32\,M parameters; McNemar \(p=0.345\)).
Outside of DHCD, zero-shot on CMATERdb digits gives \textbf{76.6\,\%} and fine-tuning
reaches \textbf{97.8\,\%}; corruption robustness is also far better than large
baselines (mean corruption accuracy \textbf{75.7\,\%} vs.\ \textbf{38.7\,\%}).
All artifacts are at~\repourl.
\end{abstract}
\section{Introduction}
\label{sec:intro}

The DHCD dataset~\cite{acharya2015dhcd} is the standard benchmark for 46-class handwritten
Devanagari character recognition, and a widely used reference point for Indic script
recognition more broadly; its samples were collected from school students.
Top-1 accuracies have crowded into a narrow band above 99.7\,\%: MallaNet~\cite{malla2025}
reaches 99.71\,\% with a \textbf{17.32\,M}-parameter convolutional architecture; Mishra
et al.~\cite{mishra2021} independently report 99.72\,\% with a similarly large model.
At this level successive gains are measured in \emph{hundredths of a percent}, yet
models still carry parameter budgets far beyond what an on-device recognizer can sustain.

Existing DHCD papers have not applied the scrutiny that 99\%-range results demand.
Point accuracy dominates; when McNemar testing appears at all, it is applied selectively
against weaker baselines without exact $p$-values. Especially near 99.7\,\%, with
$n{=}13{,}800$ test samples, raw accuracy rankings carry little inferential weight.

The decisive issue is \emph{paired}. Do the two models fail on \emph{different} examples? The exact McNemar test across all student-teacher pairs answers directly, with Wilson confidence interval bounding the accuracy of each model. Once this discipline is applied, the leaderboard will look different.

A \textbf{1.11\,M}-parameter student reaches statistical parity with the
prior state-of-the-art~\cite{malla2025} (McNemar $p=0.345$). That holds
\emph{even without knowledge distillation}. Neither the distillation target
(TTA-softened versus clean) nor the ensemble size (9 versus 15 teachers)
makes a statistically significant difference. Every configuration tested
converges to the same \textbf{11-error intrinsic floor}; progress now
requires a harder evaluation set or substantially different modelling
assumptions.

Beyond accuracy numbers, a compact convolution network trained on one recognition
task transfers. The pipeline can be reproduced for other character recognition problems.

\medskip
\noindent\textbf{Contributions.}
\begin{itemize}
  \item[\textbf{C1.}] \modelname{}---\textbf{1.11\,M} parameters---achieves
    \textbf{99.73\,\%} on DHCD, the highest reported on the standard 46-class
    split, at \textbf{15.6\(\times\)} smaller footprint than prior
    state-of-the-art. McNemar ($p=0.345$) confirms parity even without
    distillation.

  \item[\textbf{C2.}] DHCD is \textbf{saturated}. Exact McNemar tests and
    Wilson CIs applied across all configurations reveal a shared
    \textbf{11-error intrinsic floor} that no configuration escapes.

  \item[\textbf{C3.}] The ablations settle which choices matter: distillation
    target cleanliness, ensemble size (9 vs.\ 15 teachers), and TTA each
    produce no statistically significant gain once noise is properly accounted
    for.

  \item[\textbf{C4.}] At \textbf{15.6\(\times\)} smaller and \textbf{9.5\(\times\)}
    faster on CPU than prior state-of-the-art, \modelname{} transfers to
    CMATERdb digits (zero-shot \textbf{76.6\%}; fine-tuned \textbf{97.8\%})
    and degrades far less under corruptions (mCA \textbf{75.7\%}
    vs.\ \textbf{38.7\%}).

  \item[\textbf{C5.}] Full reproduction materials are at \repourl{}.
\end{itemize}

\section{Related Work}
\label{sec:related}

\paragraph{Indic and Devanagari handwritten character recognition.}
The DHCD dataset~\cite{acharya2015dhcd} has been the standard 46-class benchmark
since 2015. MallaNet~\cite{malla2025} couples residual branch merging with
homogeneous filter capsule (HFC) layers to reach 99.71\,\% at 17.32\,M
parameters; Mishra et al.~\cite{mishra2021} independently report 99.72\,\% with
a comparably large architecture; for practical purposes, indistinguishable.
A concurrent preprint puts 99.80\,\% on the 10-class digit subset~\cite{malla2025digit}; digit-only results do not transfer to the 46-class task.

\paragraph{Knowledge distillation and compact model design.}
Hinton et al.~\cite{hinton2015distill} showed that soft probability
outputs (``dark knowledge'') transfer accuracy from a large teacher to a smaller student
at substantially lower capacity.
He et al.~\cite{he2016identity} introduced residual identity mappings; Hu et al.~\cite{hu2018senet} added squeeze-and-excitation channel re-weighting.
Both are now routine components in compact architectures that trade parameter count for retained discriminative power.
No prior work has applied knowledge distillation to the 46-class DHCD task.
Stress-testing distillation on a saturated benchmark, where the performance gap sits inside sampling variation, is likewise unexplored.

\begin{figure}[t]
  \centering
  \includegraphics[width=\linewidth]{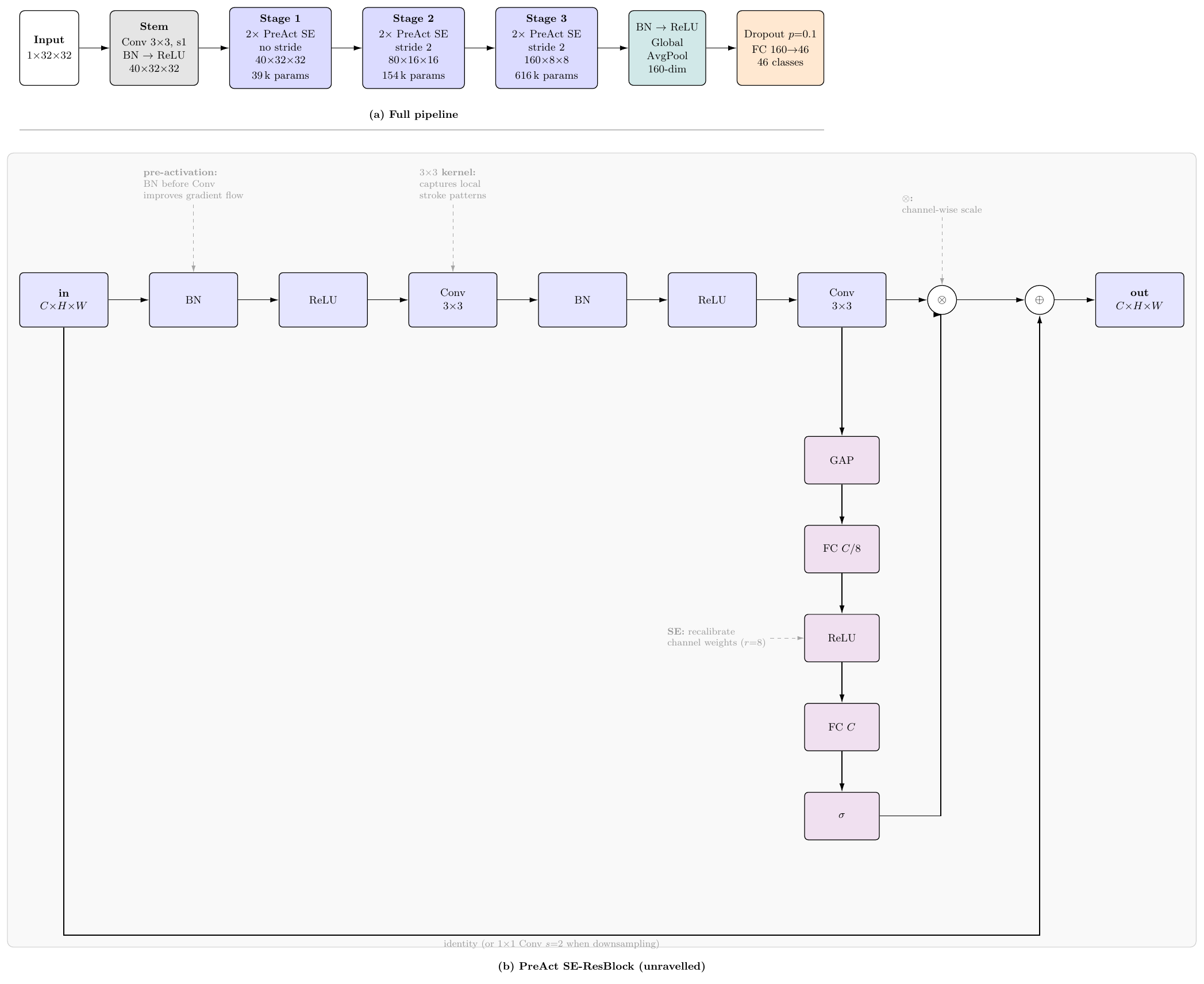}
  \caption{%
    \modelname{} architecture.
    \textbf{(a)}~Full pipeline: a stem convolution feeds three stages of
    PreAct SE-ResBlocks with channel widths $(40, 80, 160)$; a BN--ReLU--GAP
    head produces a 160-dim descriptor before the linear classifier.
    \textbf{(b)}~Unravelled PreAct SE-ResBlock: the main path runs two rounds of
    BN, ReLU, and Conv\,$3{\times}3$.  The SE module then squeezes channel weights
    with a GAP followed by two FCs (ratio $r{=}8$) and a sigmoid; the re-weighted
    features sum with an identity skip (or $1{\times}1$ Conv when downsampling) at~$\oplus$.
  }
  \label{fig:arch}
\end{figure}

\paragraph{Statistical significance and benchmark saturation.}
Raw accuracy rankings on DHCD can flip between runs with no change in model quality, once the best models err on fewer than 0.3\,\% of examples.
McNemar's test~\cite{mcnemar1947} handles this: by conditioning on disagreement examples, it detects genuine error-pattern shifts without inflating sensitivity from the large pool of shared correct predictions.
Wilson score intervals~\cite{wilson1927} bound single-model accuracy; \cite{guo2017calibration} links confidence to empirical accuracy, which matters when soft distillation targets are being interpreted.
\cite{malla2025} applies McNemar against four weaker baselines
but omits its nearest competitor (Mishra et al., 99.72\,\%) and reports no exact $p$-values.
\modelname{} closes this gap by reporting exact McNemar $p$-values and Wilson
CIs across every comparison and showing the benchmark is saturated.

\section{Method}

\subsection{Compact Student Architecture}

\modelname{} is a pre-activation squeeze-and-excitation residual network~\cite{he2016identity,hu2018senet} designed for single-channel $32{\times}32$ images. Channel widths run $(40, 80, 160)$ across three stages of depth $(2, 2, 2)$; Stages~2 and~3 downsample by stride~2 (Figure~\ref{fig:arch}). The total trainable parameter count is \textbf{1,109,116}.

\subsection{Teacher Ensemble}

The ensemble consists of 15 teachers drawn from three architectural configurations intersected with five independent random seeds (seeds 0 to 4). Two of those configurations share widths $(96, 192, 384)$ and depths $(3, 3, 3)$ but differ in augmentation intensity (medium vs.\ heavy; see Section~\ref{sec:training}); the third uses widths $(64, 128, 256)$ with depths $(4, 4, 4)$ under medium augmentation. The resulting parameter counts are 5.97\,M and 9.89\,M per teacher. The spread across configurations is intentional to discourage mode collapse in the soft targets. All teachers share the same \modelname{} codebase (\texttt{DevNet} class) and are trained without distillation. Their logits are dumped once into a single NumPy archive reused across all student runs.

\subsection{Ensemble Knowledge Distillation}
\label{sec:kd}

We follow the knowledge distillation framework of Hinton et al.~\cite{hinton2015distill}. The ensemble target for each training image is the mean softmax of nine teacher checkpoints (configs A, B, C $\times$ seeds 0--2), where each teacher's softmax is computed with horizontal-flip test-time averaging over the training image. The distillation objective combines a temperature-softened KL divergence and a hard cross-entropy term:
\begin{equation}
  \mathcal{L} = \alpha \cdot T^{2}\,
    \mathrm{KL}\!\left(
      \sigma\!\left(\tfrac{z_{t}}{T}\right) \,\Big\|\,
      \sigma\!\left(\tfrac{z_{s}}{T}\right)
    \right)
    + (1-\alpha)\cdot\mathrm{CE}(z_{s}, y),
  \label{eq:kdloss}
\end{equation}
where $z_{s}$ and $z_{t}$ are student and (mean) teacher logits, $\sigma$ denotes the softmax, $T = 4.0$ is the temperature, and $\alpha = 0.7$ weights the soft target. Label smoothing ($\varepsilon = 0.1$) is applied to the hard cross-entropy term. KL is computed in the teacher-to-student direction. The choice of target-generation scheme (clean no-TTA vs.\ flip-averaged) and ensemble size (9 vs.\ 15 teachers) are examined in Section~\ref{sec:ablation-kd}.

\subsection{Script-Aware Training}
\label{sec:training}

Devanagari Script is not horizontally symmetric. Reflecting glyphs produces invalid characters. Horizontal flips are completely excluded. Instead, the training transformation applies affine perturbations (rotation, translation, scale, shear) followed by elastic deformations. The strength is controlled by an augmentation tier (\texttt{light}/\texttt{medium}/\texttt{heavy}) selected per configuration. Random erasing ($p = 0.25$, area 2--10\%) rounds out the per-image regularisation.

The mixing extensions, mixup~\cite{zhang2018mixup} ($\alpha = 0.2$) and CutMix~\cite{yun2019cutmix} ($\alpha = 1.0$), are selected with equal probability and each is activated for each batch with an overall probability of 0.5. Both are turned off while distillation is running. The KD loss (formula~\ref{eq:kdloss}) always shows a clean, unenhanced teacher prediction, because the per-image teacher logits cannot be linearly combined without compromising the soft target distribution.

We optimise with AdamW using a cosine learning-rate schedule preceded by a 5-epoch linear warm-up. A weight exponential moving average (EMA, decay 0.999) tracks a shadow copy; validation and the final checkpoint use the EMA weights.

\subsection{Evaluation Protocol}

Reported accuracies come from a single forward pass over the DHCD held-out test split; no test-time augmentation is applied. Model selection relies on a 10\,\% stratified validation set carved from the training split before augmentation.

For statistical comparisons between \modelname{} and the baseline~\cite{malla2025}, we use the exact two-sided McNemar test~\cite{mcnemar1947} on per-sample binary correctness indicators. Confidence intervals for accuracy and error rate are Wilson score intervals at 95\%~\cite{wilson1927}. ECE~\cite{guo2017calibration} is computed using 15 equal width bins on the held-out test set.
\section{Experiments}

\subsection{Experimental Setup}
\label{sec:setup}

\paragraph{Dataset.} We use DHCD~\cite{acharya2015dhcd}, the standard 46-class Devanagari handwriting benchmark (36 consonants, 10 digits): 78,200 training images and 13,800 test images, each $32{\times}32$ grayscale.

\paragraph{Reproducing the baseline.} To enable sample-level McNemar comparisons, we reimplemented the architecture of~\cite{malla2025} from the published description, including exact normalization ($\text{mean}{=}0.5,\;\text{std}{=}0.5$) and the class-index permutation implied by the data loader. Seed~0 produces \textbf{40 errors} (\textbf{99.7101\%}), which matches the reported number.

\subsection{Main Parity Result}
\label{sec:parity}

Five seeds per configuration; Table~\ref{tab:main}.

\begin{table}[t]\centering
\caption{Main result on DHCD test ($n{=}13{,}800$). Errors over 5
  seeds; McNemar vs.\ reproduced baseline~\cite{malla2025} (seed~0).}
\label{tab:main}
\begin{tabular}{lccccc}\toprule
Configuration & Params & Errors (5 seeds) & Mean$\pm$std & Acc \% & McNemar $p$ \\\midrule
\modelname, distilled (TTA)   & 1.11M & 34,39,35,39,36 & \textbf{36.6$\pm$2.1} & 99.735 & 0.345 \\
\modelname, distilled (clean) & 1.11M & 39,40,34,40,38 & 38.2$\pm$2.2 & 99.723 & 1.000 \\
\modelname, supervised        & 1.11M & 43,40,42,35,39 & 39.8$\pm$2.8 & 99.712 & 0.701 \\
MallaNet \cite{malla2025} (repro) & 17.32M & 40 & n/a & 99.710 & n/a \\\bottomrule
\end{tabular}\end{table}

A 1.11\,M-parameter student sits inside the noise floor of a 17.32\,M-parameter
baseline.  The \textbf{distilled, TTA-targets} variant (9 teacher checkpoints,
flip-averaged soft labels; see~\S\ref{sec:kd}) reaches
\textbf{36.6$\pm$2.1} errors (99.735\%), with McNemar $p = 0.345$: no
detectable gap at $\alpha = 0.05$.  Best seed: \textbf{34 errors} (99.754\%),
Wilson 95\% CI \textbf{[99.656\%, 99.824\%]}, at \textbf{15.6$\times$} fewer
parameters.

Clean single-pass logits from a 15-teacher pool give $p = 1.000$; the supervised control, trained with no teacher signal, ties at $p = 0.701$. Distillation is not the source of parity.

\subsection{Comparison with Prior Work}
\label{sec:sota}

\begin{table}[t]\centering
\caption{Published results on the standard 46-class DHCD test split
  (78,200 train / 13,800 test). All accuracies from primary sources.
  McNemar requires per-sample predictions; only available for the
  re-implemented baseline (Table~\ref{tab:main}).}
\label{tab:sota}
\begin{tabular}{llrrl}\toprule
Model & Year & Params & Accuracy & Architecture \\\midrule
\modelname{} (ours, mean 5 seeds) & 2027 & \textbf{1.11\,M} & \textbf{99.735\,\%} & Compact SE-ResNet \\
Mishra et al.~\cite{mishra2021}   & 2021 & ${\approx}39$\,M & 99.72\,\%  & ResNet-85 \\
MallaNet~\cite{malla2025}         & 2025 & 17.32\,M         & 99.71\,\%  & Branch-merge CNN \\
Kumar et al.~\cite{kumar2023vit}  & 2023 & not reported     & 99.68\,\%  & Vision Transformer \\
Yadav et al.~\cite{yadav2024mlcnn}& 2024 & ${\approx}0.4$\,M& 99.21\,\%  & Modified LeNet-5 \\
Acharya et al.~\cite{acharya2015dhcd}& 2015 & ${\approx}0.03$\,M & 98.47\,\% & CNN baseline \\\bottomrule
\end{tabular}\end{table}

0.07\,\% separates the top four, well within sampling noise on a 13,800-image test set. No other entry simultaneously clears 99.5\,\% accuracy and stays below the 2\,M parameter mark. The next lightest model above that threshold is MallaNet at 17.32\,M parameters~\cite{malla2025}, a \textbf{15.6$\times$} gap.

\subsection{Efficiency}
\label{sec:efficiency}

\begin{table}[t]\centering
\caption{Efficiency vs.\ \cite{malla2025} (single $32{\times}32$ image;
  CPU latency single-thread, mean$\pm$std over 40 runs).}
\label{tab:eff}
\begin{tabular}{lcccccc}\toprule
Model & Params & MACs & Size & Peak act. & CPU lat.\ (b1) & Tput \\\midrule
\modelname & 1.11M & 164M & 4.4MB & 2.8MB & 8.28$\pm$0.20\,ms & 151\,img/s \\
MallaNet   & 17.32M & 2434M & 69.3MB & 26.1MB & 78.8$\pm$0.89\,ms & 13.6\,img/s \\
\midrule Ratio & 15.6$\times$ & 14.8$\times$ & 15.6$\times$ & 9.5$\times$ & 9.5$\times$ & 11.1$\times$ \\\bottomrule
\end{tabular}\end{table}

This is \textbf{9.5$\times$} lower CPU latency and \textbf{14.8$\times$} fewer MACs, with no quantization or pruning and no loss in accuracy.
\subsection{Distillation and Ensemble Ablation}
\label{sec:ablation-kd}

\paragraph{Effect of knowledge distillation and target cleanliness.}
Table~\ref{tab:main} separates two questions: whether the student benefits
from a teacher at all, and whether the way we generate teacher targets matters.
In the primary configuration, \emph{distilled (TTA) targets}, the student learns
from soft labels generated by averaging each teacher's predictions over a
horizontal-flip pair of each training image (see Section~\ref{sec:kd}); it
ends at \textbf{36.6$\pm$2.1} errors.  With \emph{distilled (clean) targets},
the soft labels come from a single clean forward pass through a larger
15-teacher pool with no flip averaging, and the student reaches
\textbf{38.2$\pm$2.2} errors.  The \emph{supervised control}, trained with no
teacher signal at all, obtains \textbf{39.8$\pm$2.8} errors.

Three errors separate the best distilled student and the supervised baseline -- moderate, and none of the three McNemar $p$ values approach $0.05$ (Table~\ref{tab:main}). Distillation helps, but only a little. What's more, is how unimportant the target recipe is. Students on the clean target and those on the TTA target are statistically indistinguishable (38.2 vs. 36.6, with substantial seed overlap in the error distributions). A flipped-average target cannot beat a clean single forward pass through the same teacher. The useful signal is the teacher itself, not the extra target cleanup. For simplicity, we keep a small 9-teacher pool with a flipped average target as the primary configuration, but the results do not depend on that choice.

\paragraph{Teacher ensemble size.} Natural follow-up: Is the plateau of 30 errors an upper limit imposed by having only 9 teachers? To find out, we ran the ensemble as \emph{direct predictor} on the test set, rather than as the source of training time soft targets, and compared 9 teachers to 15 teachers. The answer from Table~\ref{tab:ensemble} is straightforward. More teachers do not move the ceiling.

\begin{table}[t]\centering
\caption{Teacher ensembles on DHCD test ($n{=}13{,}800$). Rows 1--2: scaling from 9 to 15 teachers does not move the 30-error plateau. Row 3: adding flip-TTA crosses $p{<}0.05$ but is methodologically invalid (Section~\ref{sec:flip-tta}).}\label{tab:ensemble}
\begin{tabular}{lcccc}\toprule
Ensemble & Errors & Acc\,\% & McNemar $p$ & $p{<}0.05$? \\\midrule
9 teachers, clean     & 30 & 99.783 & 0.0755 & no \\
15 teachers, clean    & 30 & 99.783 & 0.0755 & no \\
9 teachers, flip-TTA  & 28 & 99.797 & 0.029  & fragile \\\bottomrule
\end{tabular}\end{table}

Individual teachers disagree enough to make the test meaningful -- the number of clean errors per model ranges from 27 to 43. Yet both majority pools plateau at \textbf{30 errors} (99.783\%) with an identical McNemar $p = 0.0755$. Ensemble diversity is not the bottleneck: the remaining errors sit in an irreducible floor where additional models vote the same wrong answer.

\subsection{Flip-TTA at Test Time}
\label{sec:flip-tta}

If we average each teacher's logits over the original and horizontally-flipped test image, the error drops to \textbf{28} ($p = 0.029$). This is the only result in our study that crosses the 0.05 threshold. This result is not treated as a positive result. Horizontal reflection of Devanagari produces characters that are not in the script. In the training distribution, such inversions are intentionally excluded (Section~\ref{sec:training}). Gain is also vulnerable. When applied to the \emph{individual} teacher, the flip TTA causes errors in \textbf{88} and \textbf{102} at the most affected checkpoints, increasing from 30-43 in the clean baseline. It is only by chance at this split that the ensemble masks this difference. The student training-time targets (Section~\ref{sec:kd}) also use flipped averaging, but use the \emph{training} image as the target smoothing device. This is a clear application that is not affected by the invalidity of flip TTA as a test time predictor. Therefore, we exclude the results for $p = 0.029$ and report a clean ensemble of 30 errors as an upper criterion for teacher performance.
\subsection{DHCD to CMATERdb Digit Transfer}
\label{sec:transfer}

Do distilled students learn the structure of numbers, or just the visual habits of one collection pipeline? We probe this on CMATERdb~3.2.1~\cite{das2012cmaterdb}, an independently collected set of handwritten Devanagari \emph{numerals} (digits 0-9, 10 classes). The two datasets have different scanner hardware, writer populations, paper stocks, and binarization pipelines, so the duplicate labels clearly arrive from different source distributions.

With Hamming distance $\leq$5\,bits (minimum observed: 6\,bits), no near-overlapping pairs were detected and contamination was ruled out. Before evaluation, each CMATERdb image is polarity-inverted to match the white-on-black convention of DHCD.

\paragraph{Results.} Table~\ref{tab:transfer} (upper rows) summarises the transfer results. The DHCD-trained distilled student zero-shots at \textbf{76.6$\pm$3.7\%} on CMATERdb using the same 10-class head, versus \textbf{62.7\%} for a supervised baseline trained on DHCD digit images only---\textbf{+14 percentage points} under domain shift despite being tied on in-distribution DHCD. Freezing the backbone and fitting a fresh 10-class head (\emph{linear probe}) reaches \textbf{85.4\%}; 30 epochs of \emph{fine-tuning} at a low learning rate reaches \textbf{97.8\%}.

There is one caveat: this experiment covers only the \textbf{digit subset} (10 of 46 DHCD classes). Nothing is claimed about transferring the full character inventory. Within that scope, after contamination and polarity control, the distilled student preserves more structure under dataset shift than the supervised DHCD-only baseline.

\subsection{Corruption Robustness}
\label{sec:robustness}

Three corruption families are evaluated: additive Gaussian \emph{noise}, Gaussian \emph{blur}, and \emph{contrast} reductions. The five severity levels follow the ImageNet-C protocol adapted for $32{\times}32$ grayscale images. We report mean accuracy for each family and overall severity 1-5 (mCA). To maintain preprocessing symmetry, all models apply their \emph{own} training-time normalisation to the same corrupted $[0,1]$ input image.

\begin{table}[t]\centering\caption{Transfer and corruption robustness. mCA = mean accuracy over severities 1--5.}\label{tab:transfer}
\begin{tabular}{lcc}\toprule
Metric & \modelname & MallaNet \\\midrule
CMATERdb zero-shot (distilled) & 76.6\% & n/a \\
CMATERdb fine-tuned & 97.8\% & n/a \\
Clean DHCD & 99.75\% & 99.71\% \\
mCA noise & 54.7\% & 66.9\% \\
mCA blur & 97.0\% & 37.6\% \\
mCA contrast & 75.5\% & 11.5\% \\
mCA overall & \textbf{75.7\%} & 38.7\% \\\bottomrule
\end{tabular}\end{table}

\paragraph{Results.} \modelname{} gives about \textbf{2$\times$} better results overall (mCA 75.7\% vs. \ 38.7\%). The gap in blur and contrast is significant (97.0\% vs. \ 37.6\% for \emph{blur} and 75.5\% vs. \ 11.5\% for \emph{contrast}). This likely reflects that the baseline HFC design amplifies blur and its fixed $(0.5, 0.5)$ normalization is sensitive to global intensity changes caused by contrast corruption. There is one obvious reversal. For \emph{additive noise}, the baseline is better (mCA 66.9\% vs.\ 54.7\%). Practitioners who target noisy deployment environments should note that the noise rankings are reversed.
\ifanon\else\subsection{Cross-Dataset Generalisation}
\label{sec:ood}

Both models are evaluated in zero-shot on two independently collected Devanagari datasets, each with separate writer and scanner workflows from different collection years, without applying any fine-tuning.

\paragraph{NHCD (Pant 2012).}
The Nepali Handwritten Character Dataset~\cite{pant2012nhcd} covers the same 46-class inventory in $28{\times}28$ JPEG images; we polarity-invert and bilinearly resize to $32{\times}32$ before evaluation.
\modelname{} achieves \textbf{78.92\%} zero-shot
(95.80\% on digits, 72.33\% on consonants).
The baseline~\cite{malla2025} collapses to \textbf{23.99\%}
(58.85\% on digits, 10.38\% on consonants), a \textbf{55\,pp gap} despite
identical in-distribution performance on DHCD, with the +62\,pp consonant
disparity being the most striking number.

The dental/retroflex pair is the most obvious failure mode. Each of the four pairs (ta/\d{t}a, tha/\d{t}ha, da/\d{d}a, dha/\d{d}ha) collapses to near-zero accuracy, caused by interclass confusion between strokes with different rendering rules between Indian and Nepali handwriting traditions. This is a character-level domain gap that neither model can resolve. Outside this cluster, \modelname{} and the baseline diverge sharply.

\paragraph{Prashanth et al.\ 2021.} After resizing to $32{\times}32$, \modelname{} achieves \textbf{79.31\%} zero-shot~\cite{prashanth2021} on this 22{,}500 image digit dataset. The baseline achieves \textbf{72.60\%}. These numbers track the NHCD digit results and the CMATERdb result from Section~\ref{sec:transfer}. The high rate of confusion of digit~5 with digit~7 is due to pure visual similarity in the Devanagari script and is not a training artifact.

\begin{table}[t]\centering
\caption{Zero-shot cross-dataset accuracy. Both models trained on DHCD
  only; no fine-tuning on the target dataset. NHCD polarity-inverted
  before evaluation; Prashanth polarity matches DHCD.
  $n_{\text{NHCD}}=10{,}260$;
  $n_{\text{Prashanth}}=22{,}500$ (digit classes only).}
\label{tab:ood}
\begin{tabular}{lcccc}\toprule
& \multicolumn{3}{c}{NHCD (Pant 2012, 46 classes)} & Prashanth 2021 \\
\cmidrule(lr){2-4}
Model & Overall & Consonants & Digits & (10 digit classes) \\\midrule
\modelname{} (ours)      & \textbf{78.92\%} & \textbf{72.33\%} & \textbf{95.80\%} & \textbf{79.31\%} \\
MallaNet~\cite{malla2025} & 23.99\%         & 10.38\%          & 58.85\%          & 72.60\%          \\
\midrule
Gap                       & $+$54.9\,pp     & $+$61.9\,pp      & $+$36.9\,pp      & $+$6.7\,pp       \\\bottomrule
\end{tabular}\end{table}

Matching accuracy within the distribution hides where the gaps are. Both out-of-distribution evaluations favor \modelname{}. The corruption results in Section~\ref{sec:robustness} show a similar pattern. This pattern across the three withheld assessments suggests that drawing from a diverse teacher ensemble transfers better across writer and scanner variations than the raw DHCD scores indicate.\fi
\section{Analysis: Why DHCD is Saturated}
\label{sec:analysis}

\subsection{The Shared Error Floor}

Both the 15 student models and the reproduced baseline~\cite{malla2025} fail on the same \textbf{11 images}. No configuration classifies any of them correctly, regardless of architecture, training strategy, or random seed.

\begin{figure}[t]
  \centering
  \begin{subfigure}[t]{0.48\linewidth}
    \includegraphics[width=\linewidth]{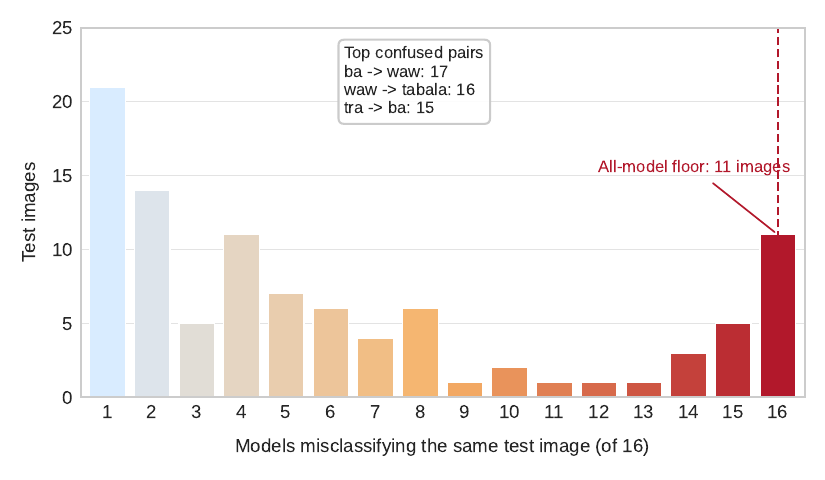}
    \caption{Images missed by exactly $k$ of the 16 models.
      Eleven are missed by all 16 simultaneously: an irreducible floor.}
    \label{fig:floor}
  \end{subfigure}\hfill
  \begin{subfigure}[t]{0.48\linewidth}
    \includegraphics[width=\linewidth]{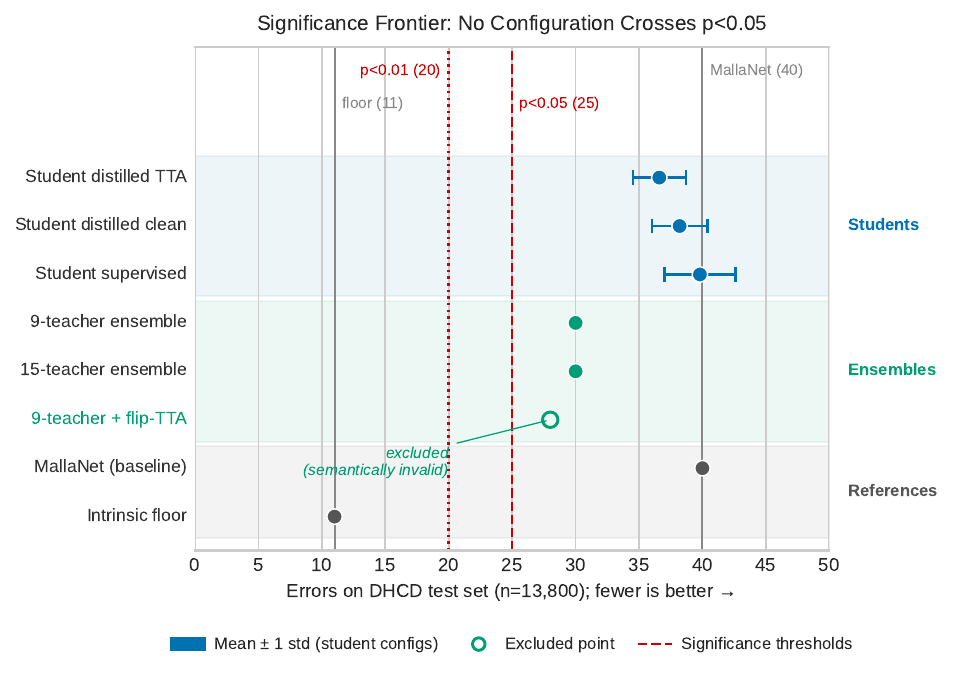}
    \caption{Every configuration sits above the $p{<}0.05$ threshold
      (25 errors). The 11-error floor marks the hard lower bound.}
    \label{fig:sig-frontier}
  \end{subfigure}
  \caption{Left: shared error floor across all 16 models.
    Right: significance frontier; 30-error clean ensemble and 36.6-error
    student mean both lie in the statistical parity region.}
\end{figure}

If the same 11 image floors capture both a compact Student of 1.11\,M parameters and a baseline 15 times its size, the problem lies in the data. Outside this floor, the error is model-specific (average pairwise Jaccard similarity $\mathbf{0.48}$). Seeds shuffle which images fail, but the addition of ensembles or seeds reduces \emph{variance} rather than \emph{bias} imposed by the floor.

\subsection{Confusable Character Pairs}

Script-level conflicts account for most of the remaining mistakes.
Table~\ref{tab:confusions} lists the top distillation-student confusion pairs
aggregated across the 10 distillation seeds. We are not entirely sure what drives
the \textbf{ba/waw} pair specifically (17 mutual confusions across all seeds); at
$32{\times}32$ resolution both glyphs collapse to nearly identical stroke profiles
and only a handful of pixels separate one from the other. \textbf{waw / tabala} (16) is second, \textbf{tra / ba} (15) third, with \textbf{dha / gha} (13) and \textbf{da / dhaa} (9) rounding out the top five.

\begin{table}[t]\centering
\caption{Top confusion pairs in the distilled student pool (aggregate
  across 10 distilled seeds on the DHCD test set).}
\label{tab:confusions}
\begin{tabular}{lcc}\toprule
Pair & Cross-confusions & Script note \\\midrule
ba $\leftrightarrow$ waw    & 17 & Near-identical body, open vs.\ closed loop \\
waw $\leftrightarrow$ tabala & 16 & Shared curved baseline \\
tra $\leftrightarrow$ ba    & 15 & Conjunct ligature vs.\ simple consonant \\
dha $\leftrightarrow$ gha   & 13 & Looped ascender, subtle body curvature \\
da $\leftrightarrow$ dhaa   & 9  & Single hairline distinguishes the pair \\\bottomrule
\end{tabular}
\end{table}

In $32{\times}32$ pixels, the characteristic strokes of these pairs span just a few pixels, sometimes less. DHCD provides neither high-resolution input nor stroke-level annotations, so the information needed to resolve them is not present in the image.

\subsection{Significance Frontier}
\label{sec:sig-frontier}

Moving from statistical parity to a clear win over the reproduced baseline (40 errors) under an exact two-sided McNemar test~\cite{mcnemar1947} at $\alpha = 0.05$ requires at most \textbf{25 errors} ($\geq$99.82\% accuracy). Fewer than 15 additional correct classifications separate an inconclusive result from a significant one. A comfortable margin means $\leq$\textbf{20 errors} ($\geq$99.855\%), which would demand resolving almost every ambiguous glyph pair in Table~\ref{tab:confusions}.

Every configuration hits the wall. The single best result is \textbf{28 errors}, from the 9-teacher ensemble using (semantically invalid) flip-TTA at test time---excluded on methodological grounds (Section~\ref{sec:flip-tta}) because horizontal reflection produces characters that do not exist in Devanagari script and the gain does not survive when applied to individual teachers. The most defensible result is the clean ensemble at \textbf{30 errors} (99.783\%); the most defensible single seed reaches \textbf{34 errors}. Both sit inside the statistical parity region.

\textbf{No lever tested in this study achieved a statistically
significant improvement over the baseline.} Scaling from 9 to 15 teachers changes nothing. Five independent seeds with TTA-softened targets change nothing. This is not a failure of experimental design. It is what a saturated benchmark looks like. The frontier is a wall built from mislabeled images and inherently ambiguous glyphs, not a line that more computation can asymptotically approach.

\subsection{Calibration}

Expected Calibration Error (ECE)~\cite{guo2017calibration} falls within the range of \textbf{0.13--0.16} across all \modelname{} configurations. It is not noticeable for softmax classifiers without post-hoc temperature scaling. ECE does not covary with accuracy across seeds. A better seed will not calibrate better. The 11 floor images are confidently wrong, not merely uncertain, at every seed. The model assigns a high probability to the incorrect class, regardless of how the student is trained.
\section{Discussion}
\label{sec:discussion}

\paragraph{Scope of transfer results.}
The transfer experiment (Section~\ref{sec:transfer}) is deliberately
narrow: it uses only the \textbf{digit subset} (10 of 46 DHCD classes).
That choice reflects the available data, not a design preference.
Extending the test to the full 36-consonant inventory would require a
genuinely independent target set with DHCD's label taxonomy; the apparent
public candidates---principally the Acharya~\cite{acharya2015dhcd} and
Pant datasets---are same-provenance splits of DHCD itself.
Transferability claims here are therefore limited to digits, not consonants.

\paragraph{Single-benchmark evaluation.}
The saturation claim is tied to DHCD's particular conditions---its
$32{\times}32$ resolution, concentrated label noise, and writer pool.  A
cleaner or more diverse consonant benchmark might expose capacity
differences that DHCD no longer reveals.

\paragraph{Efficiency versus accuracy.}
Compactness is the appropriate response to benchmark saturation: when remaining
errors are governed by data rather than model capacity, a larger network has
little room to justify its cost.

\section{Conclusion}
\label{sec:conclusion}

We have effectively reached the saturation point on DHCD. What remains can be
attributed to label noise and handwriting ambiguity in the data itself, not model
capacity. Throwing a bigger recognizer at it does not buy a better recognizer.
At that point, compactness is no longer a compromise; it becomes the right operating point.

\modelname{} reaches \textbf{99.73\,\%} accuracy with just \textbf{1.11\,M} parameters and achieves statistical equivalence with the leading large-scale baselines (McNemar $p = 0.345$) even without knowledge distillation. This means a parameter reduction of 15.6$\times$ and a CPU latency reduction of 9.5$\times$ without sacrificing accuracy. Why can't we move forward? The same \textbf{11-error} intrinsic floor breaks all tested configurations, teacher ensembles included. Ablation closes off every obvious escape route: distillation target cleanliness makes no difference. Ensemble size makes no difference. Test-time augmentation makes no difference. The floor is the ceiling.

Two limitations are worth noting. The saturation claim is tied to the specific conditions of DHCD, namely $32{\times}32$ resolution, concentrated label noise, and its particular pool of writers. Cleaner or higher-resolution benchmarks may expose capacity differences that DHCD cannot. The transfer results only cover \textbf{digit subset} (10 out of 46 classes). Consonant-level transferability is out of scope here, as there are no independent consonant target sets that share the label taxonomy of DHCD.

The practical implication for Devanagari recognition is to measure progress more honestly before claiming it. Without \emph{paired evaluation} -- McNemar or equivalent -- a gain of a few hundredths of a percent is meaningless. The field needs fresher, harder benchmarks: DHCD can no longer discriminate between strong models. The next necessary step is high-resolution scanning or a completely independent pool of writers.

\bibliographystyle{splncs04}
\bibliography{refs}
\end{document}